\documentclass{article}

     \PassOptionsToPackage{numbers, compress}{natbib}

\usepackage{hyperref}
\usepackage{wrapfig}
\AtBeginDocument{%
  \providecommand\BibTeX{{%
    \normalfont B\kern-0.5em{\scshape i\kern-0.25em b}\kern-0.8em\TeX}}}

    \usepackage[preprint]{neurips_2024}

\usepackage{graphicx}

\usepackage[utf8]{inputenc} 
\usepackage[T1]{fontenc}    
\usepackage{hyperref}       
\usepackage{url}            
\usepackage{booktabs}       
\usepackage{amsfonts}       
\usepackage{nicefrac}       
\usepackage{microtype}      
\usepackage{xcolor}      
\usepackage{enumitem}

\usepackage{quoting,xparse}
\usepackage{amsmath}
\usepackage[most]{tcolorbox}
\newcommand{\fon}[1]{\fontfamily{#1}\selectfont}  
\usepackage{quoting}
\usepackage{enumitem}
\usepackage{listings} 

\newtcolorbox[auto counter]{beliefstate}[1][]{%
fonttitle=\fon{cmr}\bfseries,
fontupper=\fon{cmr},\textrm{},
    enhanced,
    breakable,
    colback=white,
    colbacktitle=red!20!white,
attach boxed title to top center={yshift=-2mm},
    coltitle=black,
    boxrule=.6pt,
    titlerule=.2pt,
    toptitle=3pt,
    bottomtitle=3pt,
    title=Belief State
    #1}
    
\newtcolorbox{mybox}{fontupper=\fontfamily{cmr}\selectfont} 

\newtcolorbox[auto counter]{talker}[1][]{%
fonttitle=\fon{cmr}\bfseries,
fontupper=\fon{cmr},\textrm{},
    enhanced,
    breakable,
    colback=white,
    colbacktitle=grey!5!white,
attach boxed title to top center={yshift=-2mm},
    coltitle=black,
    boxrule=.6pt,
    titlerule=.2pt,
    toptitle=3pt,
    bottomtitle=3pt,
    title=Talker Agent Output,
    #1}
    
\newtcolorbox[auto counter]{reasoner}[1][]{%
fonttitle=\fon{cmr}\bfseries,
fontupper=\fon{cmr},\textrm{},
    enhanced,
    breakable,
    colback=white,
    colbacktitle=blue!20!white,
attach boxed title to top center={yshift=-2mm},
    coltitle=black,
    boxrule=.6pt,
    titlerule=.2pt,
    toptitle=3pt,
    bottomtitle=3pt,
    title=Reasoner Agent Output,
    #1}

\newtcolorbox[auto counter]{plan}[1][]{%
fonttitle=\fon{cmr}\bfseries,
fontupper=\fon{cmr},\textrm{},
    enhanced,
    breakable,
    colback=white,
    colbacktitle=green!20!white,
attach boxed title to top center={yshift=-2mm},
    coltitle=black,
    boxrule=.6pt,
    titlerule=.2pt,
    toptitle=3pt,
    bottomtitle=3pt,
    title=Plan,
    #1}

\newtcolorbox[auto counter]{user}[1][]{%
fonttitle=\fon{cmr}\bfseries,
fontupper=\fon{cmr},\textrm{},
    enhanced,
    breakable,
    colback=white,
    colbacktitle=yellow!20!white,
attach boxed title to top center={yshift=-2mm},
    coltitle=black,
    boxrule=.6pt,
    titlerule=.2pt,
    toptitle=3pt,
    bottomtitle=3pt,
    title=User,
    #1}

\title{Agents Thinking Fast and Slow: \\ A Talker-Reasoner Architecture}

\author{%
  Konstantina Christakopoulou, Shibl Mourad, Maja Matari\'c \\
  Google DeepMind\\
}

\begin{document}

\maketitle

\begin{abstract}
  Large language models have enabled agents of all kinds to interact with users through natural conversation.  Consequently, agents now have two jobs: conversing and planning/reasoning.  Their conversational responses must be informed by all available information, and their actions must help to achieve goals.  This dichotomy between conversing with the user and doing multi-step reasoning and planning can be seen as analogous to the human systems of “thinking fast and slow” as introduced by Kahneman \cite{kahneman2011thinking}.  Our approach is comprised of a "Talker" agent (System 1) that is fast and intuitive, and tasked with synthesizing the conversational response; and a "Reasoner" agent (System 2) that is slower, more deliberative, and more logical, and is tasked with multi-step reasoning and planning, calling tools, performing actions in the world, and thereby producing the new agent state. We describe the new Talker-Reasoner architecture and discuss its advantages, including modularity and decreased latency.  We ground the discussion in the context of a sleep coaching agent, in order to demonstrate real-world relevance. 
\end{abstract}

\section{Introduction}
\label{sec:intro}

Humans have the ability to do two very different kinds of thinking. On the one hand, we can form snap judgements, such as reacting to a speeding car or recognizing the emotional cues of an upset coworker. On the other hand, we can solve complicated problems, like planning a vacation and doing complex multiplications. The well-known behavioral science theory proposes that two different systems drive those abilities: the fast and intuitive System 1 and the slow and deliberative System 2 \cite{kahneman2011thinking}.
Daniel Kahneman, who introduced the theory, described the two systems for the two modes of thinking as follows:
"\emph{System 1 operates automatically and quickly, with little or no effort and no sense of voluntary control. System 2 allocates attention to the effortful mental activities that demand it, including complex computations. It represents the conscious reasoning self that has beliefs, makes choices, and decides what to think about and what to do.}"

Although difficult problems might rely more on System 2 and everyday skills more on System 1, most cognitive processes are a mix of both kinds of reasoning. 
System 1 continuously generates suggestions for System 2: impressions, intuitions, intentions, and feelings. If endorsed by System 2, impressions and intuitions form the basis of the explicit beliefs of System 2, and intentions turn into the deliberate choices of System 2. 

Many reinforcement learning (RL) problems can also benefit from a similar dual-system approach. The rapid advances in large language models (LLMs) \cite{achiam2023gpt, bubeck2023sparks, thoppilan2022lamda} have enabled artificial intelligence (AI) agents of all kinds, from AI coding buddies, to tutors and health coaches. These agents are expected to understand the complex patterns of the world via language and potentially perceive other heterogeneous multimodal signals, generating impressions, creating coherent patterns of ideas, and producing dialog (with other modalities being actively added). This resembles the coherent-seeking System 1. On the other hand, AI agents are supposed to perform complex multi-step reasoning, and make decisions that involve calling tools, actively retrieving information from external data sources, and solving complex problems. This is similar to the slower and more deliberative System 2. 

In the context of enabling agents to converse, reason and plan, in this work we consider a dual-system approach that enables those abilities through the two modes of thinking. We therefore divide the agent into two agents: a fast and intuitive Talker agent and a slower and deliberative Reasoner agent. The Talker agent focuses on generating natural and coherent conversation with the user and interacts with the environment, while the Reasoner agent focuses on performing multi-step planning, reasoning, and forming beliefs, grounded in the environment information provided by the Talker. The Talker agent, a la System 1, can access memory, priming its responses.

Similarly to the System 1 and 2 modes of thinking, the division of labor between the Talker and Reasoner agents is efficient: it minimizes effort and optimizes performance. An added benefit of this division is that the Talker can carry out the conversation, while getting more observations from the environment, without needing to wait for the slow reasoning and belief forming of the Reasoner agent. 
This is analogous to behavioral science dual-System approach, with System 1 always being on while System 2 operates at a fraction of its capacity. Similarly, the Talker is always on and interacting with the environment, while the Reasoner updates beliefs informing the Talker only when the Talker waits for it, or can read it from memory.  This division of labor works well most of the time, as the Talker is typically very good at what it does: it can automatically fetch information from memory, effectively priming its underlying model to respond well to familiar situations. However, the framework has limitations. The Talker operates with a more outdated view of the world, which has inherent biases, and can sometimes answer easier questions than the ones  asked. Also, it has little understanding of complex problem solving and planning. So, we introduce a variable allowing  the Talker to wait the Reasoner, in cases when System 2-thinking is necessary before the Talker forms its response.

To evaluate the proposed dual-system Talker-Reasoner framework, we ground our work on the real world setting of a sleep coaching agent interacting with users through dialog. We discuss success cases of this division of labor, including fast and intuitive conversations driven by the Talker and complex plans and belief states developed by the Reasoner. We also discuss cases where, similar to the dual-system thought machinery, the Reasoner (System 2) might need to override the Talker (System 1). In the AI coaching context, this could be useful when the user is requesting a complex coaching plan the Reasoner needs to finish before the Talker is able to respond.
\begin{figure}
\centering
    \includegraphics[scale=0.17]{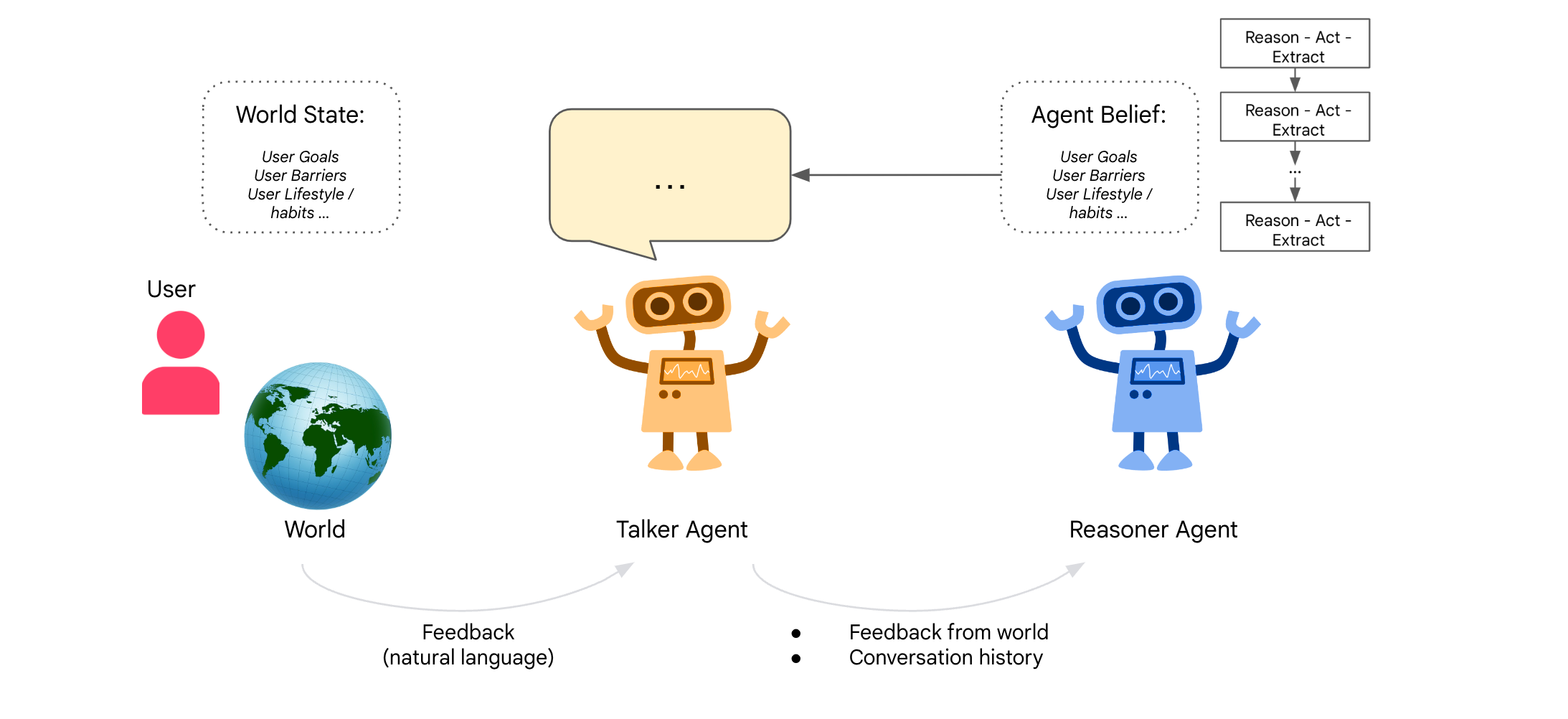}
    \label{fig:overview}
    \caption{Illustration of the proposed dual-system Talker-Reasoner approach.}
\end{figure}

\section{Related Work}
\label{subsec:related-work}
\noindent \textbf{Large Language Models for Agent Planning.} Inspired by the strong emergent capabilities of LLMs \cite{bubeck2023sparks}, such as zero-shot prompting \cite{kojima2022large}, in-context learning \cite{brown2020language}, and complex reasoning \cite{wei2022chain, zhou2024self}, research into LLM-driven agents is receiving a great deal of attention \cite{wang2023describe, zhao2024empowering, ahn2022can, shen2024hugginggpt, nakano2021webgpt, wang2023voyager, wang2024survey, sharan2023llm, brooks2024large}. The work most relevant to this paper is on \emph{text-based agents} \cite{shen2024hugginggpt, yao2022react, park2023generative}, although a great deal of work on real-world embodied agents \cite{kortenkamp2016robotic} is increasingly relevant as models become truly multimodal \cite{wang2023describe, ahn2022can, wang2023voyager, liang2023code, singh2023progprompt, szot2023large, sun2024interactive}. ReAct \cite{yao2022react} uses chain-of-thought (CoT) prompting  \cite{wei2022chain} and generates both reasoning traces and task-specific actions (e.g., tools to call) with LLMs. Reflexion \cite{shinn2024reflexion} extends ReAct with self-reflection to improve reasoning. AutoGPT \cite{yang2023auto} is a tool for automating tasks by crafting a curriculum of sub-goals for completing a high-level goal, connecting to hierarchical reasoning. 
All these works lack (i) talking while reasoning/planning and (ii) explicit belief modeling, which are crucial components in our framework. The (i) \emph{talking} aspect of our agent connects our work to prior work on \emph{natural-language based feedback agents} \cite{shinn2024reflexion, zhao2024empowering}. Similarly to \cite{zhao2024empowering}, our agent iteratively incorporates environment feedback to modify subsequent plans in a closed-loop process. However, we do not use RL to update all future plans, but instead only to adapt the cross-session plan for the single user, by augmenting the context of the in-context learned LLM with the feedback. The (ii) \emph{explicit belief modeling} aspect of our work relates to the theory of mind \cite{premack1978does, frith2003development, li2023theory} and the large body of work on world modeling \cite{forrester1971counterintuitive, ha2018world, hu2023language, zhou2023far}, as the world in our case encompasses the user, and the agent builds a proxy model of other agents (i.e., humans) to reason about their behaviors. In particular, the \texttt{Reasoner} continuously updates its belief about the user's goals, plans, barriers, motivations, in the form of a structured object or schema \cite{zhou2024self}, similarly to world and user models in past work \cite{hao2023reasoning, ha2018world, zhou2023far}. We do not yet model beliefs about beliefs (e.g., what the user thinks the agent thinks, and so on) or use world models to predict future actions; this is left for future work. 

\section{The Talker-Reasoner Agent Model}
\label{sec:formulation}
Before we introduce the dual-system Talker-Reasoner agent framework corresponding to the fast and slow thinking respectively (Section \ref{subsec:dual-agent}), we start with formalizing a \emph{single} language-based agent capable of \emph{talking} and System 1 reasoning, as well as System 2 \emph{multi-step reasoning and planning} useful for complex problem solving (Section \ref{subsec:single-agent}).  

\subsection{Single Language-Based Agent Interacting With Humans: Synergizing Talking and Extracting Beliefs With Reasoning and Planning}
\label{subsec:single-agent}
\begin{figure}
    \centering
    \includegraphics[scale=0.13]{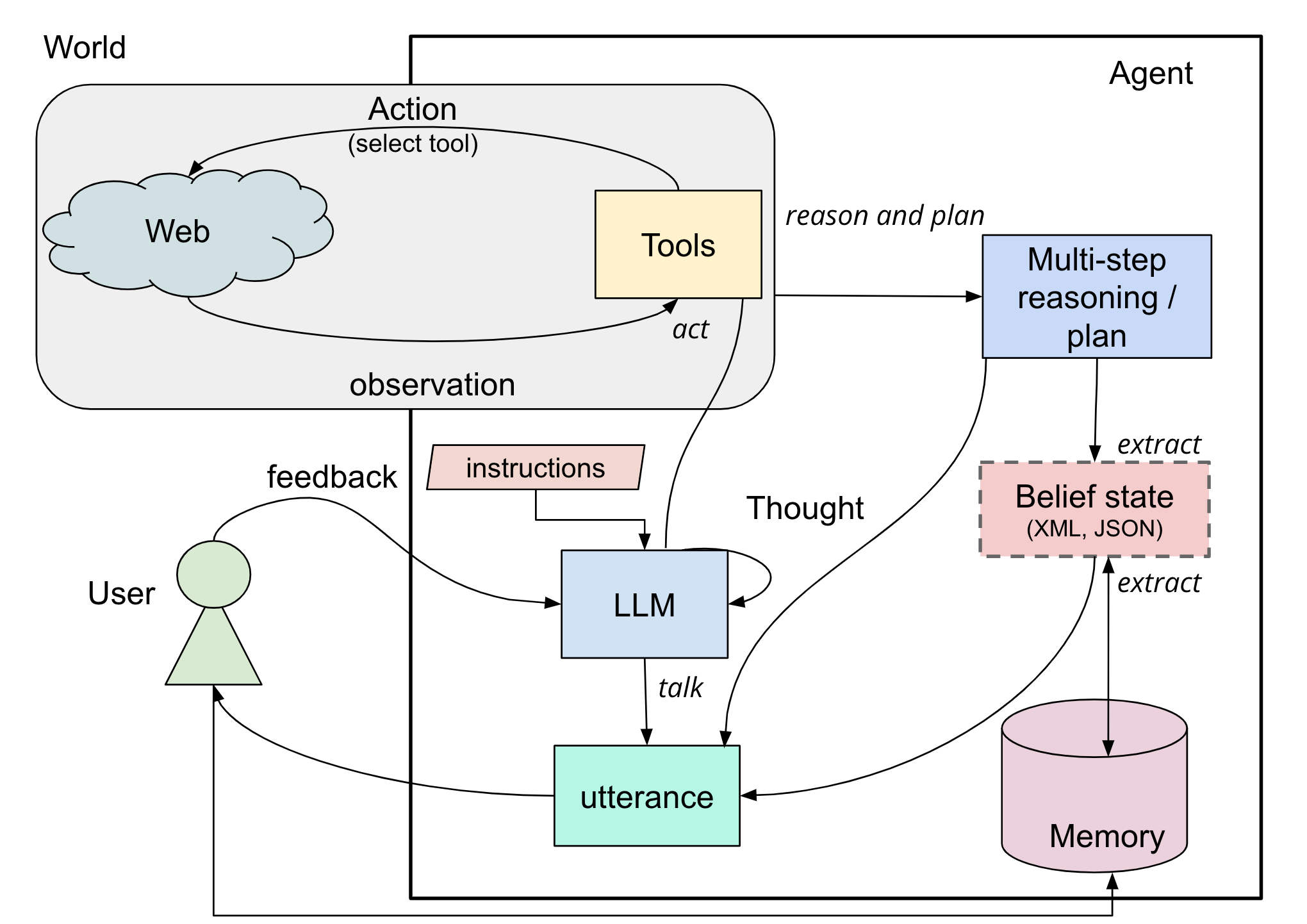}
    \caption{Single LLM-based agent that \emph{talks} and \emph{extracts} belief states while \emph{multi-step reasoning.}}
    \label{fig:overview-single-agent}
\end{figure}

Let us consider a language-based AI agent that can interact with users through natural language conversation to help them accomplish some task. The agent should be capable of multi-step reasoning and planning to be able to solve the task and also capable of generating a conversational response to the user. This paradigm of agents reasoning/planning and conversing has become more prevalent as a result of the introduction of large language models \cite{zhao2023survey, thoppilan2022lamda, achiam2023gpt, team2023gemini, touvron2023llama, sutton1999reinforcement}. We present a Reinforcement Learning (RL) formulation of this talking-and-reasoning paradigm. We also extend the paradigm to include explicit modeling of the beliefs the agent has about the user, such as the user's motivations, emotions, and goals, which guide the talking and reasoning. Figure \ref{fig:overview-single-agent} shows an overview of the overall language-based agent interacting with a user, which we will describe in detail in what follows.

We formulate the language-based agent that can reason, talk, and do explicit belief modeling in a partially-observable RL framework \cite{sutton1999reinforcement}. The agent is continuously interacting with the world $\mathcal{E}$. The world encompasses both the user the agent is interacting with, and the knowledge bases (such as the World Wide Web) that allow the agent to retrieve real-world knowledge. The agent only has a partial view of the world, thus formulating beliefs $b \in \mathcal{B}$ about the current state of the world. It can learn more about the user by interacting with them via language (future work will add other modalities). Assume that $\mathcal{L}$ represents the language space; the agent receives from the user observations $o \in \mathcal{O}$ which live in the language space $\mathcal{O} \subset \mathcal{L}$. Observations can contain both information and feedback/rewards in natural language (e.g., "I don't like this", and "Can you add something else to my plan?"). We formalize this as $\hat{\mathcal{O}} = \mathcal{O} \cup \mathcal{R}$, with observations $o \in \mathcal{L}$ and rewards $r \in \mathcal{L}$. The observations $\hat{o} \in \hat{\mathcal{O}}$ are then used to update both the agent's beliefs and the subsequent planning/reasoning performed by the agent. This can be seen as a form of online policy learning via natural language feedback and relates to natural-language based feedback agents \cite{shinn2024reflexion}. 

We now focus on the agent's actions $a \in \mathcal{A}$. The agent can (i) formulate thoughts $\tau \in \mathcal{L}$ around actions it can take, and (ii) decide which tools $a \in \mathcal{A}$ to select (e.g., APIs, engines like \texttt{SEARCH}, functions) to fetch external knowledge---this expands the space of tasks it can accomplish. By combining a series of thoughts and tools/actions, along with the results fetched via the tools, the agent can create a plan $p$ for solving a problem. Furthermore, the agent can (iii) formulate beliefs about the user (and potentially other aspects of the world); thus, another key action is \texttt{extracting} leading to a new agent belief state. The beliefs are represented as structured language objects living in XML or JSON space, with $b \in \mathcal{SL}$ \cite{zhou2023far, zhou2024self}, where structured language can be seen as a subset of the $\mathcal{L}$ language space. The belief state could encode the agent's estimate of the user's goals, needs, thoughts, sentiments, motivations, and barriers (depending on the agent context and use case), also relating this work with theory of minds \cite{li2023theory, zhou2023far}. Given the plan formed via multi-step reasoning/planning ((i) and (ii) above) and the beliefs formed ((iii) above), the agent can (iv) \texttt{talk}, i.e., provide a natural language utterance $u \in \mathcal{L}$ to the user. Putting these together, the space of actions $\mathcal{A}$, which in a classic act-only paradigm contains only the tools, is expanded to $\hat{\mathcal{A}}$. We refer to this as the {\it augmented space of actions}. The space $\hat{\mathcal{A}}$ encompasses tools $a \in \mathcal{A}$, thoughts/reasoning traces $\tau \in \mathcal{T}$ (as considered in the \texttt{ReAct} \cite{yao2022react} paradigm), and also beliefs $b$ and utterances $u$, i.e., $\hat{\mathcal{A}} = \mathcal{A} \cup \mathcal{T} \cup \mathcal{B} \cup \mathcal{U} $. While thoughts and beliefs do not affect the world and lead to no observational feedback, tools and utterances interact with the world, namely the external knowledge bases and the user, respectively.

The language-based agent chooses its augmented action according to a policy $\pi$ implemented via a large language model (LLM) with parameters $\Theta$, instructed via its prompt/constitution to follow a set of instructions $\mathcal{I} \in \mathcal{L}$. The instructions encode domain knowledge, desired behavior in interactions with the user, and the constitution \cite{bai2022constitutional} the agent is supposed to follow. The LLM implementing the policy uses the instructions, the last user's natural-language feedback/utterance $o$, along with the interaction history $\mathcal{H} \in \mathcal{L}$ in its context window $c$. Besides the context window, the agent is memory-enabled, allowing it to record all agent-user interactions along with other user information across sessions in its memory \texttt{mem}. 
That is where the agent also stores the most recent belief state $b \in \mathcal{B}$ and the current plan $p$ for how to solve the task. The agent can, at any point, retrieve relevant information from memory $x_{\texttt{mem}}$, augmenting the information in its context window. 

We formulate this in a POMDP \cite{murphy2000survey, sutton1999reinforcement}, as follows. At time step $t+1$, the language-based agent chooses actions $\hat{a} \in \hat{\mathcal{A}}$ according to a $\Theta$-parameterized LLM-based policy: 
 $ \hat{a}_{t+1} \sim \pi(\hat{a} |  c_t, x_{\texttt{mem}}; \Theta)   $
where the context $c_t = \text{Concat}(\hat{o}_t, \mathcal{H}_t, \mathcal{I})$ is the concatenation of the last user's utterance, the interaction history, and the overall instructions; $x_\texttt{mem}$ represents any information the agent might need to retrieve from memory, including the previous belief state $b$. For each user-agent interaction, when the agent needs to generate an utterance to talk to the user, there might be series of augmented actions the agent may take before it produces its response. The context captures the series of thoughts/tool selections/results obtained before it generates the action: $c_t = (\tau_1, a_1, o_1, b_1, \ldots, \tau_{t-1}, a_{t-1}, o_{t-1}, b_{t-1})$.

\begin{figure}
\centering
    \includegraphics[scale=0.13]{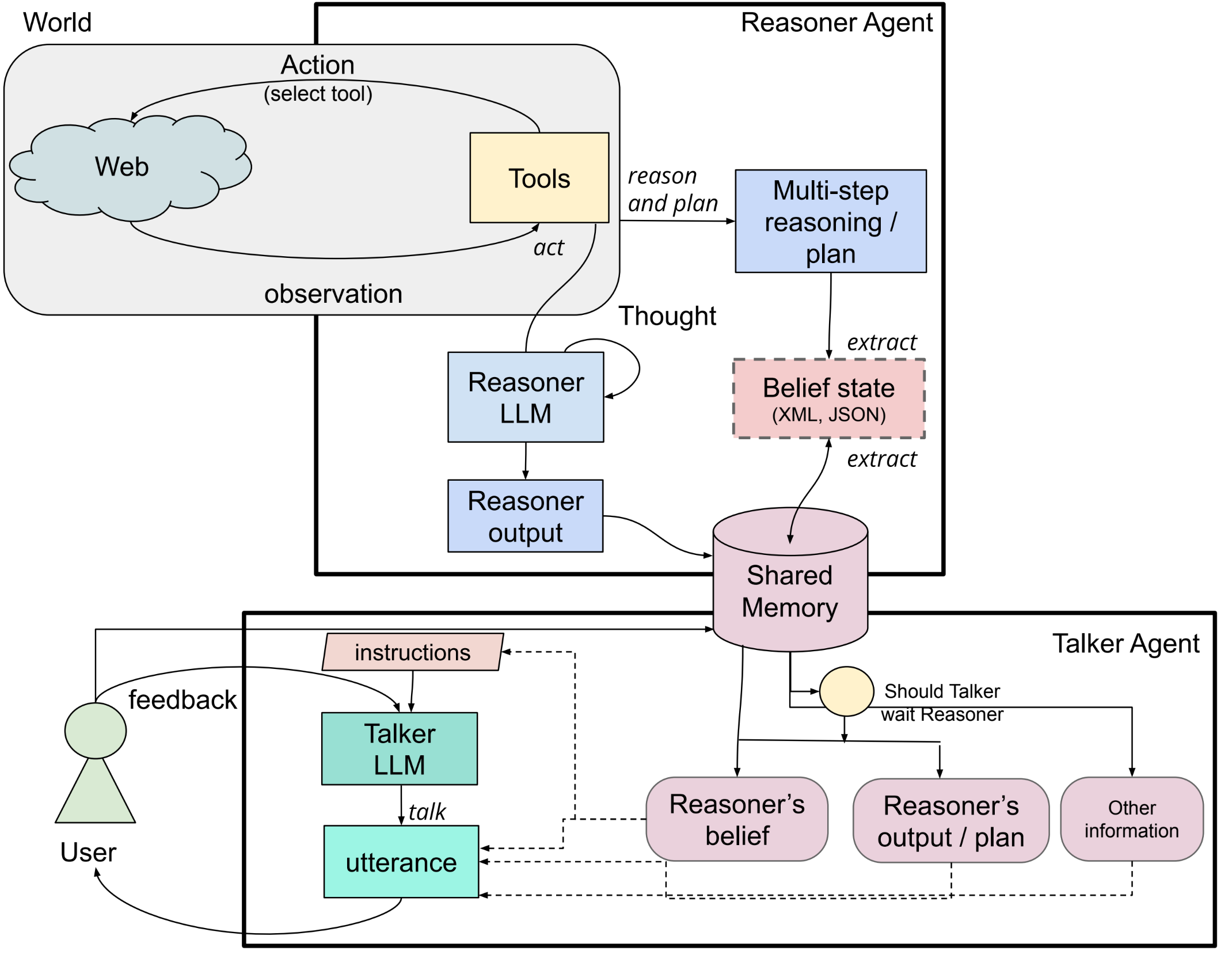}
    \caption{Diagram of Talker-Reasoner architecture.
    }
    \label{fig:talker-reasoner-architecture-diagram}
\end{figure}

\subsection{Proposed Dual-System \texttt{Talker}-\texttt{Reasoner} Agent Model}
\label{subsec:dual-agent}

So far, we have formalized an agent that can interact with users to solve tasks via its ability to do multi-step reasoning and planning, talking, and extracting beliefs about the user. However, this can be hard for a single LLM to do, as there are different requirements for talking vs. multi-step reasoning/planning and forming beliefs. In what follows, we propose the dual-system architecture, inspired by the fast and slow thinking Systems 1 and 2, respectively, consisting of: 
\begin{enumerate}[leftmargin=*]
    \item The \texttt{Talker}: The fast agent that interacts with the user via language, perceives the world, gets observations and feedback from the user, interacts with memory to prime its responses, and generates the conversational response.  
    \item The \texttt{Reasoner}: The slow and deliberative agent responsible for complex problem solving, which involves synergizing reasoning with taking actions augmenting its knowledge from the real world, such as calling tools or fetching information from external databases \cite{lewis2020retrieval}. The \texttt{Reasoner} is also responsible for making and updating beliefs that drive its decisions, and the \texttt{Talker}'s subsequent utterances. The \texttt{Reasoner} is typically  goal-conditioned, primed to solve a specific problem or goal \cite{ghosh2018learning}, and hierarchical \cite{wang2023describe}, dividing problems into sub-problems.
\end{enumerate}

As shown in Figure \ref{fig:talker-reasoner-architecture-diagram}, the main way the \texttt{Talker} (System 1) and \texttt{Reasoner} (System 2) interact is through  memory. The \texttt{Reasoner} is responsible for generating the next belief state and deriving multi-step reasoning and planning, and storing them in memory. Any time the \texttt{Talker} needs the belief, it retrieves the latest one from memory. The \texttt{Talker} therefore might operate with a delayed view of the world, as the \texttt{Reasoner} might not have had time to generate the new belief and store it in memory. However, because the \texttt{Talker} is meant to be intuitive and fast and takes into account what the user just said and the conversation history, the conversational response will still be coherent. In fact, the conversation driven by the \texttt{Talker} is processed by the \texttt{Reasoner} so that the quick impressions and responses of the \texttt{Talker} become sources of explicit beliefs and choices (plans) of the \texttt{Reasoner}. The \texttt{Talker} can also wait for the \texttt{Reasoner} before generating a response; this is equivalent to System 2 taking over and overruling System 1's impulses.

\subsubsection{The \texttt{Talker} (Thinking Fast) Agent}
The \texttt{Talker} {\it interacts with the world}, including the user, and needs to understand language and the interaction/conversation history, and be able to generate natural human-level language to do the needed \texttt{talk} action. These criteria are met by implementing the \texttt{Talker} agent with a powerful, in-context learned \cite{brown2020language} language model. Similar to System 1, the \texttt{Talker} strives for coherence, and acts as an associative machine. To ensure the coherence of the \texttt{Talker} and a good user experience, instructions $\mathcal{I} \in \mathcal{L}$ are given to the language model to follow, encoding the \texttt{Talker}'s constitution \cite{bai2022constitutional}. 

The \texttt{Talker} also {\it interacts with memory} \texttt{mem} to prime its responses with relevant information $x_{\texttt{mem}}$, including the latest beliefs that have been formed by the \texttt{Reasoner} and stored in \texttt{mem}. At every interaction with the user, the \texttt{Talker} takes the \texttt{Talk} action, and generates a conversational response, i.e., utterance $u$, conditioned on the context $c$ and the instructions $\mathcal{I}$:
\begin{equation}
\label{eq:1}
    u_{(t+1)} \sim \texttt{Talker} (u | c_{t+1}, \mathcal{I}(\cdot | b_\text{mem}); \Phi) 
\end{equation} where $\Phi$ are the parameters of the Talker.  The context $c_{t+1}$ can include the latest user utterance $\hat{o}$ which acts as both observation and natural language feedback,
the $b_{\text{mem}}$ is the latest belief produced by the \texttt{Reasoner} and stored in memory, and $\mathcal{H}_\text{mem}$ is the interaction history:
\begin{equation}
\label{eq:2}
    c_{t+1} = \text{Concat}(\hat{o}_{t+1}, b_{\text{mem}},  \mathcal{H}_\text{mem}) 
\end{equation} 
The instructions $\mathcal{I}$ the \texttt{Talker} follows can depend on the belief state. Overall, the \texttt{Talker} is supposed to be fast and conversational, minimizing latency. It is "on" whenever the user converses with the system, similar to System 1. It may use beliefs $b_{\text{mem}}$ that are not the latest $b_{t+1}$ of the \texttt{Reasoner} in order to ensure fast interactivity, meaning that the two systems may at times be decoupled.


\subsubsection{The \texttt{Reasoner} (Thinking Slow) Agent}
The \texttt{Reasoner} agent acts like System 2: it enables complex problem solving, deliberate belief forming, and choice making. 

The \texttt{Reasoner} performs \emph{multi-step reasoning and planning}, entailing series of calls to various in-context learned \cite{brown2020language} or Chain-of-Thought (CoT)-prompted language models \cite{wei2022chain}, and calls to different tools \cite{yao2022react} or databases \cite{lewis2020retrieval} for external knowledge fetching. This requires it to synergize \texttt{reasoning} (producing thoughts) and \texttt{acting} (calling tools to fetch observations), as in retrieval-augmented or tool-enhanced \texttt{ReAct}-type agents \cite{yao2022react}. The agent can develop plans (e.g., series of tools to call) and reasoning traces to solve complex tasks. It typically decomposes the problem into sub-problems in a \emph{hierarchical fashion}, and tasks each sub-problem to different modules, tools, or LLMs.
 
It also \emph{forms beliefs about the state of the world}, which can combine multiple intermediate results of multi-step reasoning, and extract from past interaction history all interesting facts about the user model in a structured language object to be stored in \texttt{mem}. This aspect of deliberate belief forming is what distinguishes the \texttt{Reasoner} from typical \texttt{ReAct}-style agents, as it includes deliberate attempt in modeling the world/human, as described in the \texttt{extract} action.

Concretely, the actions the \texttt{Reasoner} can take are: \texttt{reason}, \texttt{act}, and \texttt{extract}, each resulting in thoughts $\tau \in \mathcal{L}$, intermediate observations as a result of tool use $o$ fetching external knowledge, and beliefs $b \in \mathcal{SL}$ in the form of structured language objects. Thus, the augmented space of actions includes thoughts, tool actions, and belief extractions: $\hat{\mathcal{A}} = \mathcal{A} \cup \mathcal{T} \cup \mathcal{B}$. 
  Since the augmented action space lives in the unlimited language space, learning a policy is difficult and requires strong language priors. Thus, we implement the \texttt{Reasoner}'s policy via an in-context learned language model parameterized by $Z$. The \texttt{Reasoner} selects an augmented action:
\begin{equation}
    \hat{a} \sim \texttt{Reasoner} (b, \hat{a} | c_{\texttt{Reasoner}}; Z).
\end{equation} 
The context $c_{\texttt{Reasoner}}$ involves the interleaving of a series of $n$ thoughts, actions (e.g., calling tools like \texttt{SEARCH}), observations after these actions, and belief extractions, along with the newest observations/language feedback $\hat{o}_t$ provided by the \texttt{Talker}:
\begin{equation}
c_{\texttt{Reasoner}} = \text{Concat}(\tau_1, a_1, o_1, b_1, \ldots, \tau_n, a_n, o_n, b_n; \hat{o}_t).    
\end{equation}
When the \texttt{Reasoner} finishes its series of $n$ reasoning/planning steps, where $n$ can vary per round depending on the problem, it constructs the belief state $b_{t+1}$ as a combination of intermediate beliefs $b_1, \ldots, b_n$ and stores it in \texttt{mem}. Therefore, between two steps of user-\texttt{Talker} interaction, there are $n$ steps of slower "thinking" by the \texttt{Reasoner}.

\section{Evaluation Case Study: Sleep Coaching Agent}
\label{others}

We instantiated and validated the \texttt{Talker}-\texttt{Reasoner} dual-agent architecture in a sleep coach use case: an AI language agent interacting with users to provide help with sleeping behaviors and challenges.

\subsection{Grounding in a Real-World Scenario of AI Coaching for Sleep}
We use this real-world scenario to ground the evaluation of our dual-agent architecture. We chose AI coaching because it requires having a model of the user being coached, using sleep coaching expert knowledge to ensure scientifically-supported advice, providing a multi-step coaching plan for the user, and being conversational and empathetic much as a human coach would be. This instantiation allows us to qualitatively test the planning and reasoning capabilities of the \texttt{Reasoner} and the interactivity of the \texttt{Talker}. We also chose sleep because it is a critically important component of human health, with impact on nutrition, activity, and mental health \cite{huffington2016sleep}. The AI coach needs to continuously understand the user's needs through dialog, and to accompany the user from understanding them, to helping them set goals, to providing a multi-step plan they can follow, connecting them with resources. 


\subsection{Instantiating a \texttt{Talker}-\texttt{Reasoner} Dual-Agent Model for Sleep Coaching}

\noindent \textbf{Sleep Coaching \texttt{Talker} Agent:} We encode expert knowledge about sleep obtained from clinical experts in a set of instructions $\mathcal{I}$ that describe the agent's constitution (e.g., being empathetic, conversational, providing accurate advice) and the desired phases of sleep coaching (understanding, goal-setting, and coaching-plan) with separate instructions for each: $\mathcal{I}_{\text{understanding}}$, $\mathcal{I}_{\text{goal-setting}}$, $\mathcal{I}_{\text{coaching-plan}}$, to guide the \texttt{Talker} through the expert clinically-informed coaching process. We implemented the \texttt{Talker} via a Gemini 1.5 Flash \cite{team2023gemini} model, conditioned on the instructions, the context including the last user utterance, the interaction history, and the latest available belief state stored in \texttt{mem}, as in Equations~\ref{eq:1}, \ref{eq:2}. The model's strong language prior along with set of instructions allows it to perform complex pattern understanding and provide thoughtful conversational responses akin to System 1, as illustrated in Section \ref{subsubsec:example}.

\noindent \textbf{Sleep Coaching \texttt{Reasoner} Agent:}
The AI Coaching \texttt{Reasoner} explicitly models beliefs about the user. 
To do so, with clinician expert input, we encoded a JSON/XML schema of the belief, including fields such as sleep concern, goals, habits, barriers, and sleep environment. The \texttt{Reasoner} infers and updates the schema fields while it performs its multi-step reasoning/planning, thereby creating/updating its belief state.  We implemented a \emph{hierarchical} \cite{wang2023describe} \texttt{Reasoner}--- at every turn, it infers which of the three discrete coaching phases the user is in, and employs the corresponding of the three mini-\texttt{Reasoners}, each instantiated by a different combination of CoT-prompted Gemini 1.5 Flash models \cite{team2023gemini} along with corresponding schemas to extract.

\textbf{\texttt{Talker}-\texttt{Reasoner} Coordination:} Whether the \texttt{Talker} waits for the \texttt{Reasoner} to finish is determined by the belief---in the planning coaching phase the \texttt{Talker} waits, otherwise it does not.

\subsection{Qualitative Results}\label{subsec:qualitative}
\subsubsection{Example Conversation}
\label{subsubsec:example}
The following is an example conversation illustrating the interaction between  \texttt{Reasoner} and \texttt{Talker}.

\begin{quoting}[leftmargin=0.0cm]
\small{
    \fon{cmr}{
    \textcolor{black!95}{
    \textbf{USER}: Hey, I need your help planning re: how to create a relaxing bedtime environment. \\
    \textbf{(BELIEF in \texttt{mem})} \{updated-context: starting context, updated-title: Sleeping Coaching, coaching-phase: UNDERSTANDING\} \\
    \textbf{TALKER:} Great, let's start working on creating a relaxing bedtime environment for you :-) What's one thing in your bedroom that might be disrupting your sleep? \\
    \textbf{USER:} I think noises and light can be too distracting. Can you help create a plan for me for how to eliminate these distractions? \\
    \textbf{(BELIEF in \texttt{mem})} \emph{unchanged} \\
    \textbf{TALKER:} Absolutely! Here is a plan to eliminate distractions in your bedroom: 
    Noise:\begin{itemize}[leftmargin=*, noitemsep, topsep=0pt]
    \item Identify noise sources: Make a list of all the noise sources in your bedroom \item Block outside noise: Use blackout curtains to block light and reduce noise; Consider noise-cancelling curtains or soundproofing panels.; Try earplugs or a white noise machine. \end{itemize} Light: ...  Remember, consistency is key. Stick to this plan every night to create a peaceful and dark sleep environment. Is there anything else I can help you with? \\
    \textbf{USER:} sure, that's useful. However, I would prefer to take it more step-by-step. And also see resources of how to create a serene and relaxing atmosphere in my bedroom. \\
    \textbf{(BELIEF in \texttt{mem}):} \{user context: UserCollectedContextData(primary-sleep-concern: Noise and light distractions in the bedroom), coaching phase: CoachingPhase(value=PLANNING)\} \\
    \textbf{REASONER's PLAN:} Certainly! Let's take a more in-depth look at the first step of your plan: 
    \begin{itemize}[leftmargin=*, noitemsep, topsep=0pt]
    \itemsep0em 
    \item \emph{Step: Choose a calming color palette.} 
    Description: When selecting colors for your bedroom, opt for soothing, neutral tones like soft blues, greens, or grays. These shades have been scientifically proven to promote relaxation and tranquility.
    \begin{enumerate}[leftmargin=*]
    \itemsep0em 
    \item YouTube Video: 
    \href{https://www.youtube.com/watch?v=1GHlRJH244A}{(link)} Title: Wall Color Ideas to Create a Relaxing and Stress-free Bedroom. Channel: roomdsignTV. \underline{Reasoning:} This video from roomdsignTV provides visual examples of calming bedroom color palettes. It showcases real-life transformations that demonstrate the impact of soothing colors on creating a serene atmosphere. 
    \item YouTube Playlist: \href{https://shorturl.at/yZpMo}{(link)}   Title: Berger paints colour combination. Channel: Hresun Interiors. \underline{Reasoning:} This playlist from Hresun Interiors presents a collection of bedroom color combination ideas. It features a wide range of styles, from traditional to modern, to help you find the perfect color scheme for your bedroom.\end{enumerate} \end{itemize}
    \textbf{TALKER:} \emph{copying \texttt{Reasoner}'s response} 
    }
    }
    }
\end{quoting}

\subsubsection{Adapting planning from feedback}
The following is an example of how the \texttt{Reasoner} agent adapted its plan based on the feedback collected from user by the \texttt{Talker} agent.
\begin{quoting}[leftmargin=0.0cm]
    \small{
    \fon{cmr}{
    \textcolor{black!95}{
        \textbf{USER:} thank you, this is quite useful. Could you please add in my plan more steps around any relaxing sounds I should add to my space? \\
    \textbf{(BELIEF in \texttt{mem}):} \emph{unchanged} \\
    \textbf{PLAN by REASONER:} Absolutely! Here's an "Explore Natural Sounds" segment of your plan, along with some relevant YouTube content:
    \begin{itemize}[leftmargin=*, topsep=0pt, noitemsep]
    \itemsep0em 
        \item \emph{Explore Natural Sounds:} This segment is designed to help you create a calming and serene ambience in your space using sounds of nature. By incorporating these sounds, you can promote relaxation and reduce stress. 
        \underline{Recommended YouTube videos:} ... 
        \item \emph{Discover Healing Frequencies:} In this segment of your plan, we'll explore the world of theurapeutic frequencies, which are believed to have a profound impact on relaxation and emotional wellbeing. ... \underline{YouTube recommendations:} 
    \end{itemize}
    }}}
    \end{quoting}







\subsection{Discussion}
As mentioned in Section \ref{sec:formulation}, to minimize latency, the \texttt{Talker} uses the \emph{latest available} belief state $b$ from memory, rather than waiting for the \texttt{Reasoner} to finish its  thinking process.
The qualitative results in Section \ref{subsec:qualitative} illustrate two distinct success and failure modes of this approach:

    \noindent \emph{\textbf{"Intuitive \texttt{Talker}":}} The asynchronous approach can be effective for tasks where the \texttt{Talker} is sufficient even if it operates with an older belief state. These are typically System 1 tasks. For example, when the coaching phase is "understanding", the \texttt{Talker} can successfully carry out the conversation without the need for the \texttt{Reasoner} to finish the belief updating.
    
\noindent   \emph{\textbf{"Snap judgement \texttt{Talker}":}} However, the \texttt{Reasoner} must update its belief state before the \texttt{Talker} proceeds in complex problem-solving scenarios e.g., when the user is asking for an explicit multi-step plan or for specific resources that require tool calling. In those cases, without waiting for the \texttt{Reasoner} to finish, the Talker makes snap judgements. We can see some examples of such "snap judgement \texttt{Talker}" behavior when the belief extracted by the \texttt{Reasoner} does not yet capture the correct coaching phase, and does not fetch resources. To address this, when the \texttt{Talker} reads that the coaching phase is "planning", it is instructed to wait for the \texttt{Reasoner} to finish. This corresponds to System 2 taking over and overruling the impulses of System 1.

Finally, although there is a growing interest in AI agents performing more complex System 2 reasoning \cite{kahneman2011thinking}, we believe that our work is the first to formalize the duality of System 1 and System 2 reasoning that our \texttt{Talker}-\texttt{Reasoner} architecture offers.

\section{Conclusions}
This paper introduces the dual-system agent framework as a possible biologically-inspired architecture for foundation-model driven intelligent agents.
Inspired by the behavioral science principles behind this framework, directions for future research include deciding when not to probe the Reasoner and how to utilize it in a lower capacity most of the time, when the Talker can handle most situations. Ideally, given a user query, the Talker should automatically determine whether it requires System 2 reasoning, and therefore the Reasoner, or whether it can safely proceed with its System 1 thinking. Another direction is to extend the Talker-Reasoner architecture to multiple Reasoners, each writing belief states to different part of the memory, for different types of reasoning. 

\bibliographystyle{ACM-Reference-Format}
\bibliography{neurips_2024}


\end{document}